\documentclass[10pt,twocolumn,letterpaper]{article}

\usepackage{cvpr}
\usepackage{times}
\usepackage{epsfig}
\usepackage{graphicx}
\usepackage{amsmath}
\usepackage{amssymb}
\usepackage{color}
\usepackage{array}
\newcolumntype{P}[1]{>{\centering\arraybackslash}m{#1}}
\usepackage{multirow}
\usepackage{multicol}
\usepackage{tabu}
\usepackage{capt-of}
\usepackage{float}

\usepackage[breaklinks=true,bookmarks=false]{hyperref}

\cvprfinalcopy 

\def\cvprPaperID{5352} 

\begin{document}

\title{Few-Shot Learning with Localization in Realistic Settings}

\author{Davis Wertheimer\\
Cornell University\\
{\tt\small dww78@cornell.edu}
\and
Bharath Hariharan\\
Cornell University\\
{\tt\small bh497@cornell.edu}
}

\maketitle

\begin{abstract}
   Traditional recognition methods typically require large, artificially-balanced training classes, while few-shot learning methods are tested on artificially small ones.
   In contrast to both extremes, real world recognition problems exhibit heavy-tailed class distributions, with cluttered scenes and a mix of coarse and fine-grained class distinctions.
   We show that prior methods designed for few-shot learning do not work out of the box in these challenging conditions, based on a new ``meta-iNat'' benchmark.
   We introduce three parameter-free improvements: (a) better training procedures based on adapting cross-validation to meta-learning, 
   (b) novel architectures that localize objects using limited bounding box annotations before classification, and (c) simple parameter-free expansions of the feature space based on bilinear pooling.
   Together, these improvements double the accuracy of state-of-the-art models on meta-iNat while generalizing to prior benchmarks, complex neural architectures, and settings with substantial domain shift. 
\end{abstract}

\section{Introduction} \label{sec:intro}
\begin{figure}
    \centering
    \includegraphics[width=0.46\textwidth]{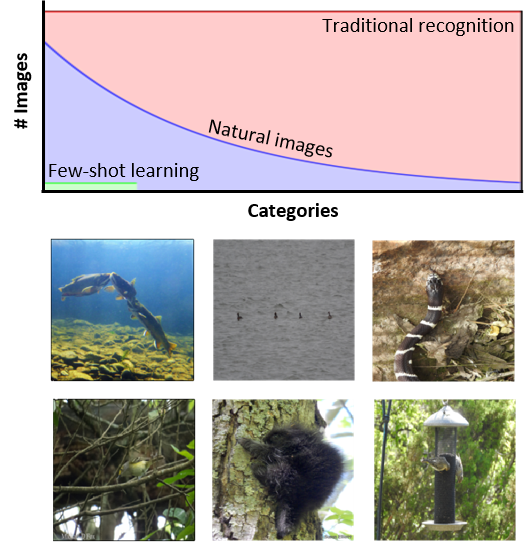}
    \caption{Discrepancies between existing benchmarks and real world problems. \textbf{Top:}  Traditional recognition benchmarks use many, equally large classes, while few-shot benchmarks use few, equally small classes. Natural problems tend to be heavy-tailed.
    \textbf{Bottom:} Clockwise from top left: relevant objects may be overlapping, tiny, occluded, underemphasized (bird is on the feeder), blurry, or simply hard to delineate \cite{inat}.}
    \label{fig:page1}
\end{figure}
Image recognition models have purportedly reached human performance on benchmarks such as ImageNet, but depend critically on \emph{large, balanced, labeled training sets} with hundreds of examples per class.
This requirement is impractical in many realistic scenarios, where concepts may be rare or have very few labeled training examples.
Furthermore, acquiring more labeled examples might require expert annotators and thus be too expensive.
This problem is exacerbated in applications (e.g., robotics) that require learning new concepts on the fly in deployment, and cannot wait for a costly offline data collection process.

These considerations have prompted research on the problem of ``few-shot'' learning: recognizing concepts from small labeled sets~\cite{FinnICML2017, BharathICCV2017, SnellNIPS2017, VinyalsNIPS2016, WangECCV2016}.
This past work builds ``learners'' that can learn to distinguish between a small number of unseen classes (often fewer than 20) based on an extremely small number of training examples (e.g. 5 per class).
However, multiple challenges plague these methods when they are applied to real-world recognition problems.

First, few-shot methods typically assume balanced datasets, and optimize the learner for an exact, often unrealistically small number of training examples per class.
In contrast, real-world problems may have highly imbalanced, heavy-tailed class distributions, with orders of magnitude more data in some classes than in others. A practical learner must therefore \emph{work equally well for all classes irrespective of the number of training examples}.
It is unclear how or even if few-shot methods can handle such an imbalance. 

Second, few-shot learning methods often assume the number of relevant concepts to be small, and as such highly distinct from each other.
In contrast, real world applications often involve thousands of classes with subtle distinctions.
These distinctions can be particularly hard to detect when natural images are cluttered or difficult to parse (Figure~\ref{fig:page1}, bottom).
Thus, the learner must also be able to \emph{make fine-grained class distinctions on cluttered natural images}.

We first evaluate prototypical networks \cite{SnellNIPS2017}, a simple yet state-of-the-art few-shot learning method, on a realistic benchmark based on the heavy-tailed class distribution and subtle class distinctions of the iNaturalist dataset \cite{inat}.
We show that prototypical networks can struggle on this challenging benchmark, confirming the intuitions above. 

We next present ways to address the challenge of heavy-tailed, fine-grained, cluttered recognition.
We introduce modifications to prototypical networks that \emph{significantly improve accuracy without increasing model complexity}.

First, to deal with heavy class imbalance, we propose a \emph{new training method based on leave-one-out cross-validation}.
This approach makes optimization easier and the learner more resilient to wider distributions of class sizes.
This technique yields a \textbf{4 point gain} in accuracy.

Second, we posit that when objects are small or scenes are cluttered, the learner may find it difficult to identify relevant objects from image-level labels alone. 
To tackle this problem, we explore \emph{new learner architectures that localize each object of interest before classifying it}. 
These learners use bounding box annotations for a tiny subset of the labelled images. 
Localization improves accuracy by \textbf{6 points}, more so when objects occupy under 40\% of the image.

Even after localizing the object, the learner may need to look for subtle distinctions between concepts.
Existing few-shot methods rely on the learning process alone to build informative feature representations.
We show that straightforward, parameter-free adjustments can significantly improve performance.
In particular, we find that \emph{the representational power of the learner can be significantly increased by leveraging bilinear pooling}~\cite{CarreiraECCV2012, new1, LinTPAMI2017}. 
While in its original formulation, bilinear pooling significantly increases the model parameter count, we show that it can be applied to prototypical networks with \emph{zero} increase.
This modification significantly improves accuracy by up to \textbf{9 points}. 

Together, these contributions \textbf{double} the accuracy of prototypical networks and other strong baselines on our challenging heavy-tailed benchmark, with negligible impact on model complexity.
Our results suggest that our proposed approach provides significant benefits over prior techniques for realistic recognition problems in the wild.

\section{Related Work}
The ideas behind our proposed techniques have broad prior support, but appear in mostly disjoint or incompatible problem settings. 
We adapt these concepts into a unified framework for recognition in real-world scenarios. 

  \noindent\textbf{Meta-learning:} Prior work on few-shot learning has mainly focused on optimizing a \emph{learner}: a function that takes a small labeled training set and an unlabeled test set as inputs, and outputs predictions on the test set. 
  This learner can be expressed as a parametric function and \emph{trained} on a dataset of ``training'' concepts so that it generalizes to new ones. Because these methods train a learner, this class of approaches is often called ``meta-learning''. Optimization may focus on the learner's parameterization \cite{feedforwardoneshot, MAML, metanets, oneshotseg}, its update schedule \cite{SNAIL, RaviLaro2017, MANN}, the generalizability of a built-in feature extractor \cite{neuralstatistician, SnellNIPS2017, VinyalsNIPS2016}, or a learned distance metric in feature space \cite{classrelevancemetrics, siameseoneshot, relation, new2}. 
  An orthogonal approach is to generate additional, synthetic data \cite{BharathICCV2017, WangCVPR2018}. 
  
  In most cases, however, few-shot classifiers \cite{feedforwardoneshot, MAML, siameseoneshot, SNAIL, RaviLaro2017, MANN, SnellNIPS2017, relation} are evaluated on one or both of only two datasets: mini-ImageNet~\cite{VinyalsNIPS2016} and Omniglot~\cite{LakeScience2015}. 
  The former presents only five classes at a time, with one or five training images per class. The latter is a handwritten character dataset, on which accuracy regularly surpasses 98\% \cite{neuralstatistician,SNAIL,metanets,relation,VinyalsNIPS2016}. 
  Some recent work has expanded the number of classes dramatically~\cite{BharathICCV2017, WangECCV2016}, but still assumes that novel classes have the same number of examples.
  These benchmarks are therefore divorced from real-world conditions, which involve difficult problems, natural images, many concepts, and varying amounts of training data \cite{deepfashion, inat, sun}. Many prior meta-learning approaches are incompatible with these settings.
  Notably, Wang et al~\cite{WangNIPS2017} design an approach to heavy-tailed problems based on knowledge transfer from common to rare classes.
  Their approach is orthogonal to our improvements.
  
  \noindent\textbf{Heavy-tailed datasets:} Heavy-tailed class distributions are common in the real world. MS-COCO \cite{mscoco}, the SUN database \cite{sun}, DeepFashion \cite{deepfashion}, MINC \cite{minc}, and Places \cite{places} are all examples where an order of magnitude separates the number of images in the most versus the least common classes. MINC and Places are especially noteworthy because they are explicitly designed to \emph{narrow} this gap in data availability \cite{minc, places}, yet display heavy class imbalance anyway. Despite this trend, standard recognition benchmarks like ImageNet \cite{imgnet}, CIFAR-10, and CIFAR-100 \cite{cifar} heavily curate their data to ensure that classes remain nicely balanced and easily separable. The mini-ImageNet and Omniglot few-shot benchmarks encode class balance explicitly, as do other proposed few-shot benchmarks~\cite{BharathICCV2017, WangECCV2016, WangCVPR2018}.
  
  
  \noindent\textbf{Improving feature space:} It is well known that higher-order expansions of feature space can raise the expressive power of hand-designed feature extractors \cite{vlad, fv}. Recent work has shown that similar techniques \cite{CarreiraECCV2012, new1, LinTPAMI2017}, learnable generalizations of these techniques \cite{circproject, deepfried}, and efficient approximations to these techniques \cite{compactbili, walshbili} also improve the performance of convolutional networks. The improvement is especially large in fine-grained classification settings, such as facial recognition \cite{covafacial, bilifacial, LinTPAMI2017}. However, using the resulting expanded feature space requires parameter-heavy models, even in the few-shot setting \cite{new2}. We adapt bilinear pooling \cite{LinTPAMI2017} as a truly parameter-free expansion, which no longer risks overfitting to small datasets.

  \noindent\textbf{Localization:} A close relationship exists between localization and recognition. Networks trained solely on image-level, classification-based losses nevertheless learn to localize objects of interest \cite{locfree1, locfree2}. These learned localizations can act as useful data annotation, including for the original recognition task \cite{residualattention, twolevel, locfree2}. Very difficult problems, however, may require expensive ground truth annotations to begin bootstrapping. Fortunately, a very small set of annotations can be sufficient to predict the rest \cite{oneshotseg}. Semi-supervised localization further improves when image-level category labels are provided \cite{segprop2, segprop1}. Since each can bootstrap from the other, combining recognition and localization may prove a particularly effective remedy for data scarcity.

\section{Problem Setup and Benchmark}

\begin{figure}
    \centering
    \includegraphics[width=0.48\textwidth]{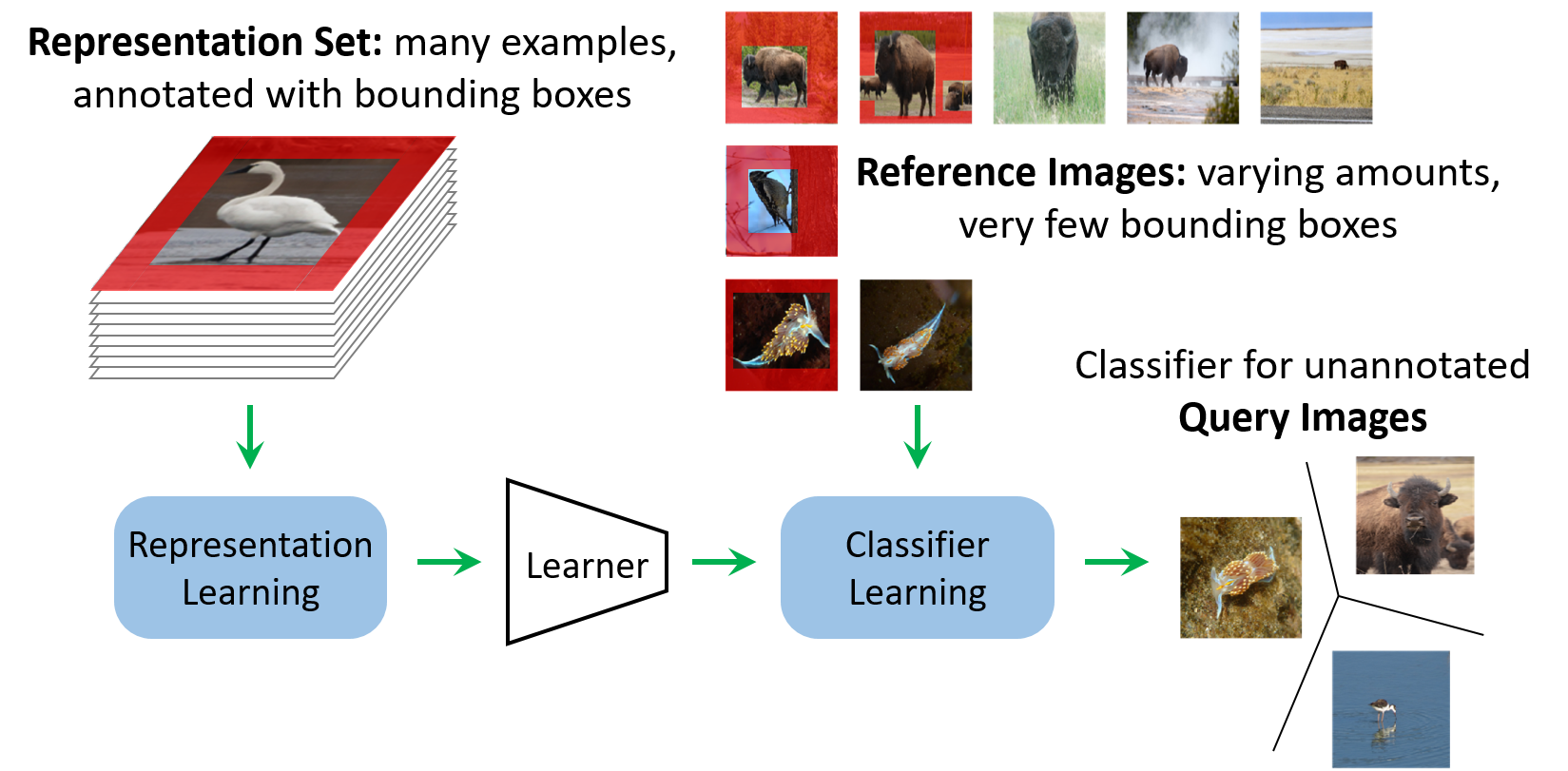}
    \caption{Our real-world learning benchmark. Initially, many images are available with bounding box annotations. The learner must then adapt to new classes using varying but limited amounts of data, with very few bounding boxes. At test time in the wild, there are no annotations.}
    \label{fig:page2}
\end{figure}

  Our goal is to build \emph{learners}, systems that can automatically learn new concepts under challenging real-world conditions, with heavy-tailed distributions of classes and subtle class distinctions.
  Each learner may have tunable parameters or hyperparameters.
  As in prior work, these parameters are learned on a ``representation set'' of concepts (``base classes'' in~\cite{BharathICCV2017}) with many training examples (see Fig. \ref{fig:page2}).
  
  Once trained, the learner must generalize to a disjoint ``evaluation set'' of novel categories.
  The evaluation set is split into a small collection of labeled ``reference images'' and a larger set of unlabeled ``query images''.
  The learner may use the reference images to define the new set of categories, estimate new parameters for those categories (e.g. a linear classifier) and/or fine-tune its feature representations. 
    
  Final accuracy is reported on the unannotated query images. 
  We report top-1 and top-5 accuracy, both as a mean over images and over the categories of the evaluation set. 
  The latter metric penalizes classifiers that focus on large categories while ignoring smaller ones. 
  
  Two approaches to the above problem act as illustrative examples.
    A traditional transfer learning approach is to train a softmax classifier on the representation set. On the evaluation set, the fully-connected layer is replaced by a new version with the appropriate number of categories, and fine-tuned on reference images. Query images form the test set. 
    Meta-learning approaches, such as prototypical networks, train a parametric learner on tiny datasets sampled from the representation set, teaching the learner to adapt to novel tiny datasets. The learner processes the evaluation set in a single pass, with reference images forming the training set and query images forming the test set.

\noindent \textbf{Object location annotations:} \label{ssec:annotation}
  As discussed in Section~\ref{sec:intro}, a key challenge in real-world recognition problems is finding relevant objects in cluttered scenes. 
  Small sets of image-level class labels may be insufficient. 
  We therefore provide bounding boxes for \emph{a small fraction ($\le 10\%$) of the reference images in the evaluation set}.
  Note that with extremal point clicks, these annotations are cheap to acquire in practice~\cite{extremal}.
  We fully annotate the representation set, as such datasets tend to be heavily curated in the real world  (Fig. \ref{fig:page2}).

  \subsection{Benchmark Implementation}
    We now convert this problem setup into a benchmark that accurately evaluates learners on real-world heavy-tailed problems.
    For this, the evaluation set must satisfy three key properties. 
    First, as in many real-world problems, training sets should be heavily imbalanced, with orders of magnitude difference between rare and common classes.
    Yet the number of examples per class must be neither unnecessarily small (e.g. fewer than 10), nor unrealistically large (e.g. more than 200).
    Second, in contrast to past few-shot learning benchmarks that use five classes at a time \cite{LakeScience2015, VinyalsNIPS2016}, there should be many (e.g. at least 20) categories in the evaluation set, with coarse- and fine-grained distinctions, as in the real world.
    Third, images must be realistically challenging, with clutter and small regions of interest.
    
\begin{figure}
    \centering
    \includegraphics[width=0.46\textwidth]{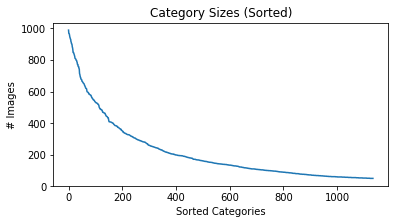}
    \caption{Category sizes in meta-iNat}
    \label{fig:cat_sizes}
\end{figure}

    We implement our benchmark using the iNat2017 dataset \cite{inat}, an organically collected, crowdsourced compendium of living organisms, with fine- and coarse-grained species distinctions, a heavy-tailed class size distribution, and bounding box annotations for a significant subset.
    Of the appropriately-sized categories with bounding boxes, 80\% are randomly assigned to the representation set, and the rest to the evaluation set.
    Within the evaluation set, 20\% of images are reference images and the rest are query images, for an overall split of 80/4/16\% representation, reference, and query.
    We propose this ``meta-iNat'' dataset as a realistic, heavy-tailed, fine-grained benchmark for meta-learning algorithms.
    Meta-iNat contains 1,135 animal species, the distribution for which can be found in Fig. \ref{fig:cat_sizes}. 

    While all images in meta-iNat have bounding box annotations, only 10\% are allowed during evaluation (see Section \ref{ssec:annotation}). We run ten trials on the evaluation set with a different collection of annotated reference images in each trial.

\section{Approach}
We build upon prototypical networks \cite{SnellNIPS2017} (Section \ref{ssec:protonet}) by introducing three light-weight and parameter-free improvements. Batch folding (Section \ref{ssec:batchfold}) improves gradients during training and helps the learner generalize to large classes. Few-shot localization (Section \ref{ssec:localization}) teaches the learner to localize an object before classifying it. Covariance pooling (Section \ref{ssec:covapool}) greatly increases the expressive power of prototype vectors without affecting the underlying network architecture. In addition to being parameter-free, these techniques are mutually compatible and mutually beneficial.

  \subsection{Prototypical Networks}
  \label{ssec:protonet}
    We briefly review prototypical networks \cite{SnellNIPS2017}. 
    Prototypical networks are a learner architecture designed to learn novel classes using few training examples. 
    The learner uses a feature extractor to embed labeled reference and unlabeled query images in feature space.
    Reference image embeddings are averaged within each class to generate a ``prototype'' vector for that class. 
    Predictions are made on query embeddings based on L2 proximity to each class prototype. 
    
    Training a prototypical network amounts to setting the parameters of the feature extractor, since classification is non-parametric.
    The prototypical network is trained on the representation set by sampling \emph{small} datasets of reference and query images.
    These are passed through the network to get class probabilities for the query images.
    Cross entropy loss on query images is then minimized.
    Through this training, the network learns a feature extractor that produces good prototypes from limited reference images.

  \subsection{Batch Folding} \label{ssec:batchfold}
    Batch folding is motivated by the fact that during training, every image in a batch is either a reference or a query image, but never both. 
    While reference images learn to \emph{form} good class centroids, conditioned on the other contributors, query images \emph{gravitate toward} the correct centroid and \emph{away} from others. 
    Gradients for both are necessary for learning, but every image gets only one, so prototypical weight updates are noisy. 
    
    This reference/query distinction also limits the number of reference images a network can handle.
    For a prototype network to work on common classes as well as rare ones, it must be trained with a larger number of reference images~\cite{SnellNIPS2017}. 
    Increasing the reference images per batch, however, requires either increasing the batch size, which runs into memory constraints, or decreasing the number of queries, producing noisier query gradients.
    Thus the original prototypical networks are \emph{designed} for rare classes.
    
    As an alternative, we propose to use leave-one-out cross-validation within each batch, abandoning the hard reference/query split. The entire batch is treated as reference images, and the contribution of each image is subtracted (``folded'') out from its corresponding prototype whenever it acts as a query. 
    Each image thus gets a combined, cleaner gradient, acting as both a reference and a query. 
    Furthermore, the number of query / reference images can be as high as the batch size / one less.
    The result is stable training with large reference sets without violating memory constraints.
    We call this approach batch folding.
    
    \noindent\textbf{Procedure:}
        Let $n$ be the number of classes and $p$ the number of images per class in a batch.
        Denote by $v_{i,j}$ the feature vector of the $i$-th image in the $j$-th category.
        Let $c_j = \frac{\sum_i v_{i,j}}{p}$ be the centroid of the $j$-th class.
        To make predictions for the $i$-th image in the $j$-th category, the network uses the following class prototypes:
        \begin{align} \label{eq:fold}
            c_1,\-\ c_2,\-\ ...\-\ c_{j-1},\-\ \frac{p}{p-1}(c_j-\frac{v_{i,j}}{p}),\-\ c_{j+1},\-\ ...\-\ c_n
        \end{align}
    
    \noindent\textbf{Overhead:}
        Batch folding is efficiently parallelizable using tensor broadcasting. 
        The necessary broadcast operations are built-in to most machine learning libraries, including NumPy \cite{numpydocs}, PyTorch \cite{pytorchdocs}, and TensorFlow \cite{tensorflowdocs}. 
        
        Note also that standard prototypical network prediction already involves calculating the L2 distance between every centroid and every query image embedding. This has the same asymptotic cost as calculating (1) for every image, so long as 
        query set size $n_{query} \approx n_{total}$. Generally this is true \cite{SnellNIPS2017}. The overhead of batch folding also tends to be dominated by earlier convolutional layers.

  \subsection{Localization} \label{ssec:localization}
    Image-level labels are less informative when the object of interest is small and the scene cluttered, since it is unclear what part of the image the label refers to.
    Given many, sufficiently different training images, the machine eventually figures out the region of interest~\cite{locfree2}.
    But with only a few images and image-level labels, distinguishing relevant features from distractors becomes highly difficult.
    
    For these reasons, isolating the region of interest (on \emph{both} reference and query images) should make classification significantly easier. 
    We consider two possible approaches. 
    In \textbf{unsupervised} localization, the learner internally develops a category-agnostic ``foregroundness'' model on the representation set.
    \textbf{Few-shot} localization uses reference image bounding boxes on the \emph{evaluation set} for such localization. 

    \noindent\textbf{Procedure:}
\begin{figure*}
    \centering
    \includegraphics[width=\textwidth]{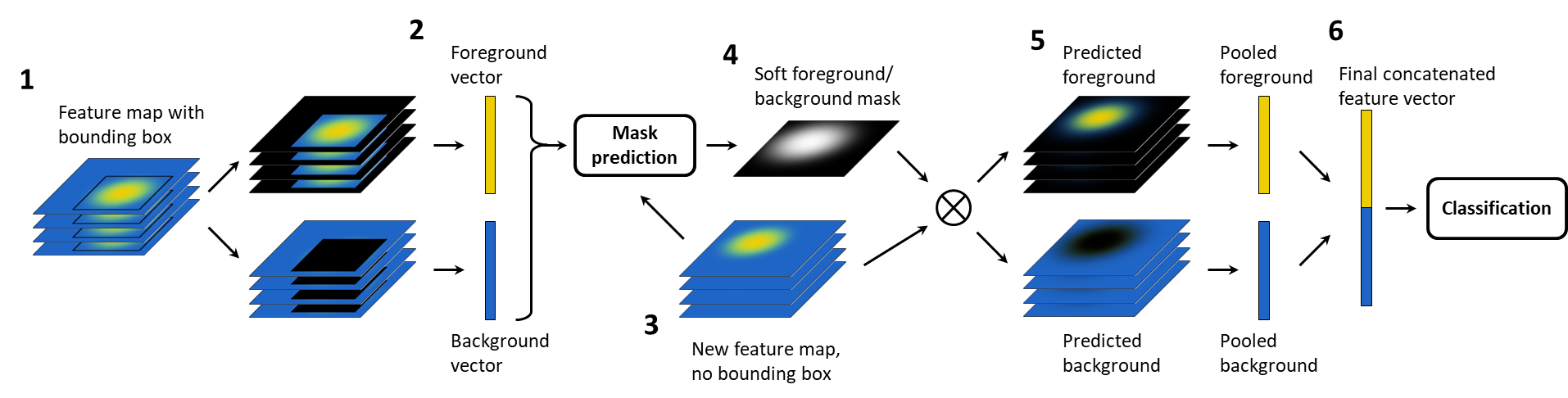}
    \caption{Few-shot localization. Provided bounding boxes mask off foreground and background regions (1), which are averaged to produce foreground and background feature vectors (2). Pixel features on new feature maps (3) are classified as foreground or background based on distance from those vectors (4). The predicted mask separates foreground and background regions (5), which are average pooled independently and concatenated (6).
    Unsupervised localization learns the foreground/background vectors as parameters, and begins at (3).}
    \label{fig:fbsep}
\end{figure*}
        In both approaches, the localizer is a sub-module that classifies each location in the final $10\times 10$ feature map as ``foreground'' or ``background''.
        This prediction is calculated as a softmax over each pixel embedding's negative L2 proximity to a foreground vector and to a background vector.
        In unsupervised localization, these vectors are learned parameters optimized on the representation set. In few-shot localization, the localizer gets a few reference images annotated with bounding boxes. We use these boxes as figure/ground masks, and average all the foreground pixel embeddings to produce the foreground vector.
        The background vector is computed similarly.
    
        The output of the localizer is a soft foreground / background mask. Multiplying the feature map with its mask (and inverse mask) produces foreground and background maps, which are average-pooled then concatenated. This double-length feature vector is used to form prototypes and perform classification.
        Fig. \ref{fig:fbsep} provides a visual explanation.

\begin{figure}
    \centering
    \includegraphics[width=0.46\textwidth]{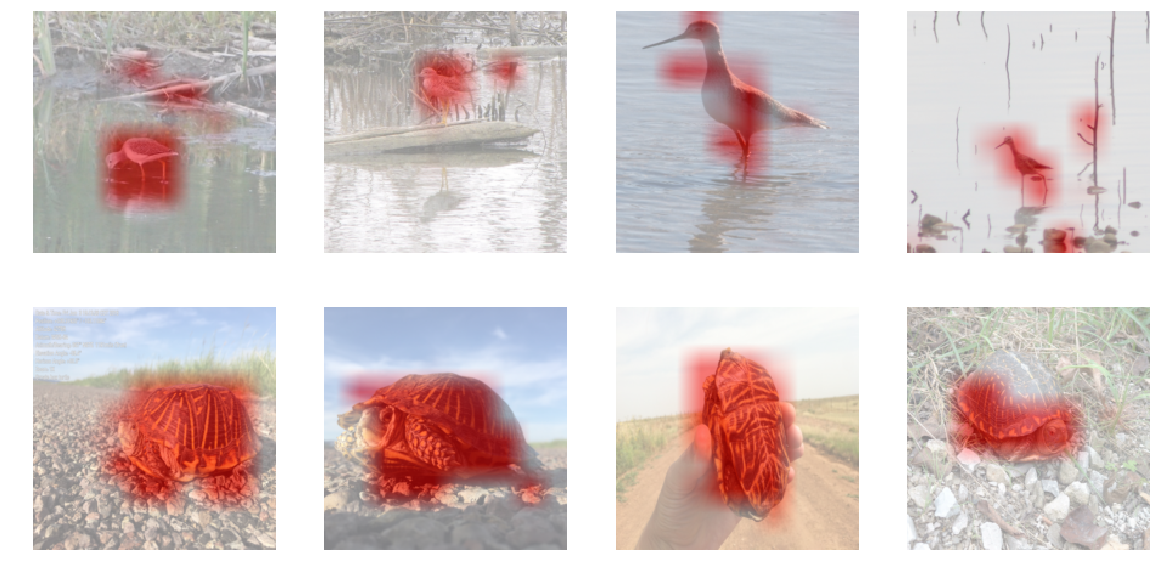}
    \caption{Example outputs of the few-shot localizer. The leftmost image provides the foreground and background centroids for each row. The network learns without supervision or dedicated parameters to isolate (mostly) appropriate regions of interest.}
    \label{fig:mask_visualizations}
\end{figure}

    \noindent\textbf{Training:}
        Both localization approaches are trainable end-to-end, so we train them within the classification problem.
        We use no additional supervisory loss; localizers are trained only to be useful for classification.
        Despite this, the outputs are visually quite good. Examples are given in Fig. \ref{fig:mask_visualizations}. 
        
        When a few-shot localizer is trained with batch folding, an additional round of folding during localization is required. Each image's contribution is removed from the foreground and background vectors. Otherwise, each image `sees' its own ground truth bounding box during localization, preventing generalization to unannotated images. 

  \subsection{Covariance Pooling} \label{ssec:covapool}
    For hard classification problems, methods such as bilinear pooling~\cite{LinTPAMI2017}, fisher vectors~\cite{fv} and others~\cite{covafacial, compactbili, new1} can be used to expand the feature space and increase expressive power.
    Unfortunately, a traditional learning framework uses these expanded representations as input to linear classifiers, fully-connected softmax layers, or multilayer networks \cite{bilifacial, walshbili, LinTPAMI2017, new2}, dramatically increasing parameters and making the model prone to catastrophic overfitting. 
    
   However, these techniques can be adapted to prototypical networks without any parameter increase.
   We use bilinear pooling~\cite{LinTPAMI2017},\footnote{
   Similar techniques have been called second-order pooling~\cite{CarreiraECCV2012}, higher-order pooling~\cite{new1}, and covariance descriptors~\cite{covardescript} in the literature.}
   which improves fine-grained classification performance and generalizes many hand-designed feature descriptors such as VLAD \cite{vlad}, Fisher vectors \cite{fv}, and Bag-of-Visual-Words \cite{bovw}. 
   This approach takes two feature maps (from e.g. a two-stream convolutional network) and computes the cross-covariance between them, by performing a pixel-wise outer product before average-pooling. 
    In our localization models, the predicted foreground and background maps act as the two streams. 
    Otherwise, we use the outer product of the feature map with itself. 
    Both versions perform signed square-root normalization, as in bilinear pooling, but do not project to the unit sphere, as this heavily constrains the prototype prediction space. 
    
    It is worth emphasizing that this expansion adds no parameters. 
    Unlike prior models, all improvement in performance comes from increased feature expressiveness, not from increased network capacity.
    To emphasize this distinction, we call this version covariance pooling.

\section{Experiments}
We first present overall results on the meta-iNat benchmark (Table \ref{tab:baselines}).
We analyze localizer behavior, and then generalization to larger networks, tasks with domain shift, and the original mini-ImageNet. 
We use 4-layer convolutional learners closely mimicking prototypical networks \cite{SnellNIPS2017}, plus average-pooling (see supplementary). 

  \subsection{Meta-iNat}
    \begin{table}
    \centering
    \resizebox{.49\textwidth}{!}{
    \scriptsize
    \setlength\tabcolsep{5pt}
    \setlength\extrarowheight{1pt}
    \hskip-.02\textwidth
    \begin{tabular} {  l  P{1.23cm}  P{1.23cm}  P{1.23cm}  P{1.23cm}}
        \hline
        \-\ & \multicolumn{2}{c}{\textbf{Top-1 Accuracy}} & \multicolumn{2}{c}{\textbf{Top-5 Accuracy}} \\
        
        \textbf{Model} & Mean & Per-Class & Mean & Per-Class \\
        \hline
        Softmax & 13.35$\pm$.24 & 6.55$\pm$.19 & 34.46$\pm$.30 & 20.05$\pm$.30 \\
       
        Reweighted Softmax & 6.92$\pm$.19 & 7.88$\pm$.16 & 21.94$\pm$.31 & 22.53$\pm$.29 \\
        
        Resampled Softmax & 1.54$\pm$.06 & .99$\pm$.02 & 3.77$\pm$.01 & 2.75$\pm$.03 \\
     
        Transfer Learning & 17.39$\pm$.24 & 17.61$\pm$.10 & 41.03$\pm$.25 & 40.81$\pm$.27 \\
     
        PN & 16.07$\pm$.19 & 17.55$\pm$.19 & 42.1$\pm$.21 & 41.98$\pm$.18 \\

        \hline

        PN+BF & 20.04$\pm$.04 & 20.81$\pm$.08 & 47.86$\pm$.31 & 46.57$\pm$.23 \\

        PN+BF+fsL* & 26.25$\pm$.05 & 26.29$\pm$.04 & 55.43$\pm$.09 & 53.01$\pm$.08 \\

        PN+BF+usL & 28.75$\pm$.13 & 28.39$\pm$.15 & 57.90$\pm$.24 & 55.27$\pm$.37 \\

        PN+BF+usL+CP & 32.74$\pm$.13 & 30.52$\pm$.13 & 61.32$\pm$.14 & 56.62$\pm$.16 \\

        PN+BF+fsL+CP* & \textbf{35.52$\pm$.05} & \textbf{31.69$\pm$.06} & \textbf{63.76$\pm$.09} & \textbf{57.33$\pm$.10} \\
        \hline
    \end{tabular}
    }
    \vspace{.8mm}
    \caption{Results on the meta-iNat benchmark, with 95\% confidence intervals from 4 trials. PN is a prototypical network, BF is batch folding, fsL and usL are few-shot and unsupervised localization, and CP is covariance pooling. *Results are averaged over 10 runs of 4 trials, annotations randomly sampled per-run.}
    \label{tab:baselines}
    \end{table}
  
  \noindent\textbf{Baseline results:}
    Standard softmax classifiers trained from scratch on the evaluation set's reference images perform poorly, especially on rare classes.
    Upweighting rare classes during training improves the per-class accuracy only slightly. 
    Oversampling the rare classes causes catastrophic overfitting.    
    A second baseline is transfer learning: we train the same network on the representation set, but replace and re-train the final linear layer on the evaluation set, using class weights. This approach works significantly better than training from scratch, attaining 17.6\% per-class accuracy.
    
    As our third baseline, prototypical networks trained on the meta-iNat representation set easily outperform the models trained from scratch, and are comparable to transfer learning without any need for label reweighting.
    This suggests that prototypical networks are inherently class-balanced, but provide no additional advantages over transfer learning in this heavy-tailed setting.
    
    \noindent\textbf{Batch folding:}
\begin{figure}
    \centering
    \includegraphics[width=0.46\textwidth]{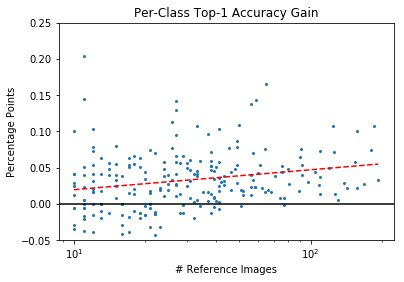}
    \caption{Batch folding improves accuracy for all class sizes in expectation, but particularly helps with large ones ($r^2=.05$)}
    \label{fig:foldgain}
\end{figure}
    A prototypical network trained with batch folding outperforms all baselines by almost a \textbf{3-point margin}. 
    The per-class accuracy gain as a function of class size is plotted in Fig. \ref{fig:foldgain}. 
    We see gains across the board, suggesting that batch folding does provide higher-quality gradients.
    At the same time, by incorporating more reference images during training, batch folding helps models generalize to larger classes: the positive slope of the best-fit line suggests that large classes benefit more from batch folding, though not at the expense of small ones.

    \noindent\textbf{Localization:}
    Incorporating few-shot localization leads to another significant boost in performance, about \textbf{6 percentage points}.
    Note that 10\% of reference images are annotated, only 1 to 20 images per category. 
    This relatively cheap annotation has an outsized impact on performance. 
    
    Interestingly, unsupervised localization provides a larger gain, about \textbf{8 percentage points}.
    We posit that few-shot localization underperforms its counterpart because it uses bounding boxes, a very coarse form of segmentation. Bounding boxes may include a significant amount of background, hurting the separation of foreground from background.
    Indeed, we find that when provided bounding boxes are large (e.g. occupying the full image), the few-shot localizer is unable to localize correctly.
    
\begin{figure}
    \centering
    \includegraphics[width=0.46\textwidth]{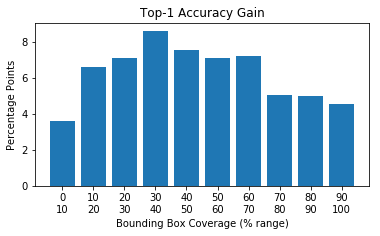}
    \caption{Few-shot localization is more helpful on images with small regions of interest}
    \label{fig:localgain}
\end{figure}
    As hypothesized, localization particularly helps when objects are small, and bounding boxes cover less than half the image (Fig. \ref{fig:localgain}). The decrease in gain for tiny objects is not entirely surprising - classification is inherently harder when the relevant object contains only a few pixels. 
    
\begin{figure}
    \centering
    \includegraphics[width=0.46\textwidth]{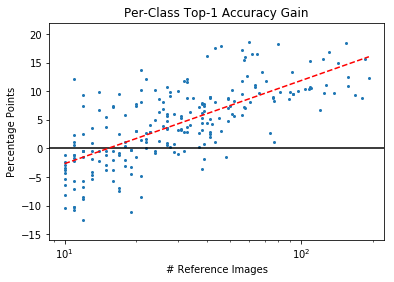}
    \caption{Covariance pooling improves performance on large classes, at the expense of some small ones ($r^2=.50$)}
    \label{fig:covagain}
\end{figure}
    \noindent\textbf{Covariance pooling:}
    Accuracy improves yet again with covariance pooling, yielding a \textbf{4 point} gain over unsupervised localization and \textbf{9 points} over few-shot localization.
    Notably, covariance pooling causes class balance to break: large categories benefit disproportionately (Fig. \ref{fig:covagain}). We hypothesize that the high dimensionality of covariance space is responsible. Small categories do not have enough reference images to span the space, so centroid quality suffers.  

    Unsupervised localization does not interact well with covariance pooling, perhaps because covariance space is too high-dimensional for reference images to span during training. Thus the learned foreground and background vectors may overfit to a particular manifold on the representation set.
    Few-shot localization, which calculates these vectors dynamically, does not have this problem.
    We conclude that both localization techniques are useful for different settings.
    
    Using all three techniques, top-1 accuracy is doubled from the baseline prototypical network. 
    The best performer uses \textbf{batch folding, few-shot localization} and \textbf{covariance pooling}.
    An ablation study is provided in supplementary. 
    
  \subsection{Analyzing Few-Shot Localizer Behavior}
  
  \begin{table}
    \centering
    \resizebox{0.99\linewidth}{!}{
    \scriptsize
    \setlength\tabcolsep{5pt}
    \setlength\extrarowheight{1pt}
    \hskip-.01\textwidth
    \begin{tabular} { l c c c }
        \hline
                \textbf{Localizer} & \textbf{\% annotation}  & \textbf{Mean acc.} & \textbf{Per-Class acc.}  \\
         \hline
        Untrained & 10\% & 19.74$\pm$.03 & 20.42$\pm$.06  \\
        No Gradient & 10\% & 22.77$\pm$.23 & 22.86$\pm$.18  \\
        \hline
        & Supercategory & 23.67$\pm$.79 & 24.08$\pm$.66  \\
        &1\%  & 25.85$\pm$.11 & 25.96$\pm$.09  \\
        Jointly trained & 4\%  & 26.17$\pm$.08 & 26.22$\pm$.06  \\
        & 16\%  & \textbf{26.28$\pm$.05} & \textbf{26.3$\pm$.04}  \\
        & 64\% & 26.21$\pm$.04 & 26.25$\pm$.03  \\
        \hline
    \end{tabular}%
    }
    \vspace{.8mm}
    \caption{Top-1 accuracy with 95\% confidence intervals for few-shot localization models as annotations increase, with comparison to baseline localizers. All models use batch folding.}
    \label{tab:local}
    \end{table}

    We next evaluate the number of bounding box annotations needed for good classification accuracy.
    As shown in Table~\ref{tab:local}, performance saturates at 16\% bounding box availability, but even at 1\% (amounting to one box per class), performance decreases only slightly.
    This scarcity can go further: categories in meta-iNat are grouped into nine supercategories, so we also try using one box per supercategory, nine total. 
    Accuracy does drop significantly, but is still better than models that do not localize. Thus localization can lead to real accuracy gains using hardly any annotations at all, to our knowledge a first-of-its-kind finding. 
    
    \noindent\textbf{Joint training:} Although the few-shot localizer never receives direct training supervision, it must still be learned jointly with the classifier. Table~\ref{tab:local} also compares localizers that are not jointly trained.
    Applying few-shot localization to a network trained without it leads to a drop in performance (``Untrained''). 
    Training the network to use localization, but preventing backpropagation through the localizer itself, also leads to a drop in performance (``No Gradient''). 
    Localization thus provides a useful training signal, but must itself be trained with the classifier for maximum benefit.

  \subsection{Generalization}
    We evaluate our models on three new settings. To test generalization over domain shift, we create a second split of meta-iNat based on supercategories. 
    To test generalization to other network architectures, we evaluate our techniques on meta-iNat using more powerful, pretrained ResNet architectures. 
    Finally, these techniques are tested on mini-ImageNet, using evaluation methods from prior literature. With some expected caveats for mini-ImageNet, our results generalize extremely well to all settings. 
  
    \noindent\textbf{Supercategory meta-iNat:}
        We wish to evaluate our results in settings where transfer learning is more difficult, and switching from the representation set to the evaluation set involves substantial domain shift. To that end we construct a new version of meta-iNat, which we call Supercategory meta-iNat. Rather than assign categories to the representation and evaluation sets randomly, we instead split by supercategory. Insects and arachnids (354 total) form the evaluation set, and everything else (birds, fish, mammals, reptiles, etc.) is the representation set. Training and evaluation are performed as before, with results in Table \ref{tab:nonR}. 
    \begin{table}
    \centering
    \resizebox{.49\textwidth}{!}{
    \scriptsize
    \setlength\tabcolsep{5pt}
    \setlength\extrarowheight{1pt}
    \hskip-.02\textwidth
    \begin{tabular}{ l P{1.23cm} P{1.23cm} P{1.23cm} P{1.23cm}}
        \hline
        \-\ & \multicolumn{2}{c}{\textbf{Top-1 Accuracy}} & \multicolumn{2}{c}{\textbf{Top-5 Accuracy}} \\
        \textbf{Model} & Mean & Per-Class & Mean & Per-Class \\
        \hline
        Reweighted Softmax & 4.59$\pm$.21 & 5.38$\pm$.22 & 15.95$\pm$.58 & 16.57$\pm$.53 \\
        Transfer Learning & 6.34$\pm$.23 & 6.19$\pm$.14 & 18.89$\pm$.48 & 17.86$\pm$.49 \\
        PN & 5.33$\pm$.18 & 6.31$\pm$.18 & 17.41$\pm$.45 & 18.32$\pm$.27 \\
        \hline
        PN+BF & 7.29$\pm$.11 & 8.24$\pm$.13 & 22.09$\pm$.35 & 22.53$\pm$.37 \\
        PN+BF+fsL* & 11.69$\pm$.06 & 12.38$\pm$.07 & 30.64$\pm$.11 & 29.86$\pm$.09 \\
        PN+BF+usL & 12.46$\pm$.59 & 12.95$\pm$.51 & 32.28$\pm$1.1 & 31.18$\pm$.95 \\
        PN+BF+usL+CP & 17.65$\pm$.21 & 16.72$\pm$.18 & 40.16$\pm$.26 & 36.19$\pm$.48 \\
        PN+BF+fsL+CP* & \textbf{20.02$\pm$.13} & \textbf{17.32$\pm$.09} & \textbf{43.45$\pm$.20} & \textbf{36.65$\pm$.15} \\
        \hline
    \end{tabular}
    }
    \vspace{.8mm}
    \caption{Results on the Supercategory meta-iNat benchmark, with 95\% confidence intervals. Models are as in Table \ref{tab:baselines}.}
    \label{tab:nonR}
    \end{table}
        
        Transfer learning on Supercategory meta-iNat is much harder than in the original setting. Scores are uniformly lower across the board. However, overall trends remain exactly the same. 
        Batch folding outperforms standard prototypical networks and transfer learning baselines by \textbf{2 points}. Few-shot and unsupervised localization lead to similar, substantial accuracy gains (\textbf{4 points}). Covariance pooling also improves (\textbf{5 points}), but again causes mean accuracy to outstrip per-class accuracy. Unsupervised localization underperforms few-shot localization when using covariance pooling, so we remove it from future tests.

    \noindent\textbf{ResNet-50:}
        While batch folding, few-shot localization, and covariance pooling lead to substantial improvement on meta-iNat, accuracy is still low. For more powerful models, these improvements might disappear. To test this, we replace the bottom two prototypical network layers with a frozen ResNet-50 pretrained on ImageNet. Details can be found in supplementary. 
        Results are presented in Table \ref{tab:resnet50}. 
    \begin{table}
    \centering
    \resizebox{.49\textwidth}{!}{
    \scriptsize
    \setlength\tabcolsep{5pt}
    \setlength\extrarowheight{1pt}
    \hskip-.02\textwidth
    \begin{tabular} { l P{1.23cm} P{1.23cm} P{1.23cm} P{1.23cm}}
        \hline
        \-\ & \multicolumn{2}{c}{\textbf{Top-1 Accuracy}} & \multicolumn{2}{c}{\textbf{Top-5 Accuracy}} \\
        \textbf{Model} & Mean & Per-Class & Mean & Per-Class \\
        \hline
        Transfer Learn (top) & 19.27$\pm$.17 & 18.72$\pm$.20 & 44.02$\pm$.30 & 41.2$\pm$.36 \\
        Transfer Learn (full) & 22.52$\pm$.58 & 18.22$\pm$.40 & 48.16$\pm$.60 & 40.38$\pm$.48 \\
        PN & 35.35$\pm$.24 & 35.59$\pm$.11 & 67.82$\pm$.13 & 66.33$\pm$.19 \\
        \hline
        PN+BF & 37.36$\pm$.15 & 36.73$\pm$.12 & 69.25$\pm$.16 & 67.03$\pm$.15 \\
        PN+BF+fsL* & 46.2$\pm$.04 & 44.43$\pm$.08 & 75.87$\pm$.04 & \textbf{73.26$\pm$.06} \\
        PN+BF+fsL+CP* & \textbf{51.25$\pm$.13} & \textbf{46.04$\pm$.13} & \textbf{77.5$\pm$.06} & 72.14$\pm$.05 \\
        \hline
    \end{tabular}
    }
    \vspace{.8mm}
    \caption{Results on meta-iNat using ResNet50 features, with 95\% confidence intervals. Transfer Learning (top) adjusts unfrozen upper layers on reference images, while (full) fine-tunes the entire network. Other models are as in Table \ref{tab:baselines}.}
    \label{tab:resnet50}
    \end{table}
        
        Using the pretrained ResNet-50 model, it is possible to perform transfer learning directly from ImageNet to the meta-iNat evaluation set. Freezing the ResNet, and training just the top two layers on reference images, works poorly given the power of the model. Fine-tuning the entire network on reference images works slightly better, but decreases per-class accuracy. Freezing the ResNet and training the top layers as a prototypical network improves top-1 accuracy by \textbf{13 percentage points}. Batch folding, few-shot localization, and covariance pooling provide another \textbf{16 points}. We conclude that these techniques are helpful for large neural architectures as well as small ones. 

    \noindent\textbf{Mini-ImageNet:}
    \begin{table}
    \centering
    \scriptsize
    \setlength\tabcolsep{5pt}
    \setlength\extrarowheight{1pt}
    \begin{tabular} { l P{2cm} P{2cm} }
        \hline
        \textbf{Model} & \textbf{5-shot Accuracy} & \textbf{1-shot Accuracy} \\
        \hline
        PN & 65.76$\pm$.29 & 49.97$\pm$.30 \\
        PN+BF & 65.2$\pm$.29 & 47.67$\pm$.31 \\
        PN+fsL & 67.85$\pm$.29 & \textbf{51.1$\pm$.3} \\
        PN+fsL+CP & \textbf{69.45$\pm$.28} & 49.64$\pm$.31 \\
        \hline
    \end{tabular}
    \vspace{.8mm}
    \caption{Five-class accuracy on mini-ImageNet with 95\% confidence intervals over 10 dataset passes. ``Shot'' refers to number of reference images. Models are as in Table \ref{tab:baselines}.}
    \label{tab:miniimgnet}
    \end{table}
        Batch folding, few-shot localization, and covariance pooling improve accuracy on large evaluation sets with long-tailed class distributions. 
        To see if these techniques still help with the original, smaller few-shot learning problem, we construct a mock mini-ImageNet dataset with similar statistics but annotated with bounding boxes. Performance of prototypical networks on our dataset is similar to the published figures~\cite{SnellNIPS2017}. Table \ref{tab:miniimgnet} shows the results.
    
\begin{figure}
    \centering
    \includegraphics[width=0.46\textwidth]{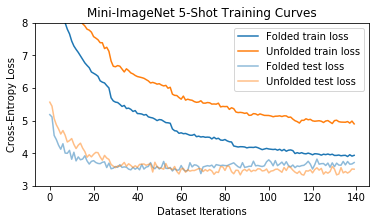}
    \caption{Batch folding causes overfitting on the smaller mini-ImageNet representation set. Models are trained on twenty categories but tested on five, so test loss is lower than training loss.}
    \label{fig:overfitting}
\end{figure}

        An immediate departure from prior results is the fact that batch folding hurts performance. 
        Batch folding does indeed result in better training and lower training loss, but overfits because the representation set is smaller (Fig. \ref{fig:overfitting}).
        
        Few-shot localization and covariance pooling make modest but real improvements when five reference images are provided (``five shot''). There is little discernible effect on single-reference (``one-shot'') performance, perhaps because with few reference images and a small set of classes, a more expressive feature space is unnecessary.
        Nevertheless, the small improvements do suggest that few-shot localization and covariance pooling generalize to few-shot learning. 
        
\section{Conclusion}
In this paper, we have shown that past work on classical or few-shot balanced benchmarks fails to generalize to realistic heavy-tailed classification problems. 
We show that parameter-free localization from limited bounding box annotations, and improvements to training and representation, provide large gains beyond those previously observed in data abundant settings. 
Ours is but a first step in addressing broader questions of class balance and data scarcity.

\noindent \textbf{Acknowledgements} This work was partly funded by a grant from Aricent.

{\small
\bibliographystyle{ieee}
\bibliography{egbib}
}

\newpage
\onecolumn

\begin{center}
    \LARGE
    \textbf{Supplementary Materials}
\end{center}

\noindent \textbf{Note}: Code for the models and main results of the paper are available at \href{https://github.com/daviswer/fewshotlocal}{\textcolor{blue}{https://github.com/daviswer/fewshotlocal}}. The full set of supplementary materials, including additional results, localizer visualizations and behavior, and the exact category assignments in meta-iNat by species, can be found \href{https://daviswer.github.io/files/cvpr2019supp.pdf}{\textcolor{blue}{here}}.

\section{Ablation Study}
    Top-1 and top-5 mean and per-class accuracies are given below. Three-digit model names indicate the presence or absence of batch folding, localization, and covariance pooling, in that order. For example, `101' indicates a model with batch folding and covariance pooling, but no localization. Because two versions of localization exist, we use `0' to indicate no localization, `1' for few-shot localization, and `2' for unsupervised localization. A `*' indicates a model presented in the main paper. 
    
    \begin{table}[H]
    \centering
    \resizebox{.6\textwidth}{!}{
    \scriptsize
    \setlength\tabcolsep{5pt}
    \setlength\extrarowheight{1pt}
    \hskip-.02\textwidth
    \begin{tabular} {  l  P{1.5cm}  P{1.5cm}  P{1.5cm}  P{1.5cm}}
        \hline
        \-\ & \multicolumn{2}{c}{\textbf{Top-1 Accuracy}} & \multicolumn{2}{c}{\textbf{Top-5 Accuracy}} \\
        
        \textbf{Model} & Mean & Per-Class & Mean & Per-Class \\
        \hline
        000* & $16.07\pm.19$ & $17.55\pm.19$ & $42.10\pm.21$ & $41.98\pm.18$ \\
        100* & $20.04\pm.04$ & $20.81\pm.08$ & $47.86\pm.31$ & $46.57\pm.23$ \\
        010 & $21.55\pm.09$ & $22.37\pm.08$ & $50.20\pm.08$ & $48.70\pm.08$ \\
        001 & $24.31\pm.48$ & $24.64\pm.39$ & $53.29\pm.76$ & $51.09\pm.63$ \\
        020 & $24.32\pm.15$ & $24.68\pm.81$ & $53.16\pm.89$ & $51.44\pm.94$ \\
        021 & $25.60\pm1.02$ & $25.51\pm.90$ & $54.66\pm1.07$ & $52.04\pm.91$ \\
        110* & $26.25\pm.05$ & $26.29\pm.04$ & $55.43\pm.09$ & $53.01\pm.08$ \\
        011 & $27.46\pm.15$ & $26.39\pm.14$ & $56.37\pm.17$ & $52.37\pm.17$ \\
        120* & $28.75\pm.13$ & $28.39\pm.15$ & $57.90\pm.24$ & $55.27\pm.37$ \\
        101 & $31.06\pm.61$ & $29.07\pm.53$ & $60.41\pm.67$ & $55.50\pm.62$ \\
        121* & $32.74\pm.13$ & $30.52\pm.13$ & $61.32\pm.14$ & $56.62\pm.16$ \\
        111* & $35.52\pm.05$ & $31.69\pm.06$ & $63.76\pm.09$ & $57.33\pm.10$ \\
        \hline
    \end{tabular}
    }
    \end{table}
    
    Between any two models with different numbers of features, the one with more always outperforms the one with less, regardless of combination. Thus batch folding, localization, and covariance pooling consistently and independently improve performance on meta-iNat.

\section{Network Architectures}
    Our learner architectures mimic the original prototypical networks. Networks contain four 64-channel $3\times 3$ convolutional layers, with Batchnorm, ReLU, and $2\times 2$ max-pooling in between. This is followed by Batchnorm and $10\times 10$ global average-pooling, for a 64-dimensional feature vector. Unlike prototypical networks, which flatten the feature map to a 6400-dimensional vector, we use average pooling to eliminate spatial priors on uncentered, uncropped images. Softmax classifier models use a fully-connected layer on top of the feature vector; prototype classifiers use the feature vector directly. Localization and covariance pooling models replace average-pooling layers with the appropriate algorithm(s). 
    
    Models based on ResNet50 use the first two stages of a ResNet50 model, pretrained on ImageNet, to produce $28\times 28$ feature maps with 512 channels. Learned layers consist of $2\times 2$ max-pooling, followed by $3\times 3$ convolution with 128 channels, Batchnorm, ReLU, a second $3\times 3$ convolution with 64 channels, and Batchnorm. The resulting $14\times 14$ feature maps are localized, average pooled, or covariance pooled as appropriate. As in the original setting, softmax classifiers use an additional fully-connected layer, while prototype architectures use the pooled embeddings directly.

\section{Training and Implementation Details}
\subsection{Training}
    We train meta-iNat models via SGD with Adam, using an initial learning rate of $10^{-3}$ and dividing by two every epoch. Each epoch consists of 10 passes over the representation set, with five epochs in total. For mini-ImageNet, we lengthen each epoch to 28 passes, to account for the smaller representation set. 
    
    In both settings we unit-normalize input color channels. We use random horizontal flipping but no other data augmentation. Because bounding boxes are downsized to very small resolutions (e.g. $10\times 10$), we use $4\times 4$ average-pooling before downsampling, as a fast approximation for anti-aliasing.

\subsection{Batch Sampling}
    While mini-ImageNet uses random batch sampling with replacement during training and testing, meta-iNat uses a different sampling procedure for each. During each training iteration, classes are sampled without replacement: so long as classes have sufficient remaining images to create a batch, they are sampled proportionately to the number of available images they contain. This ensures that differently-sized classes occur at constant, representative rates throughout the entire sampling process. Images are then selected from the sampled classes, and those images are subsequently unavailable for further training until the next pass over the dataset.
    
    During testing, we wish to evaluate classification accuracy for relatively large numbers of classes and images. For sufficiently high values, it becomes impossible to process the sampled datasets as single batches in computer memory. Category sizes also vary so widely that it no longer makes sense to use constant sample sizes. Rather than attempt to evaluate a given network on a large number of large and complicated datasets, we instead impose a single reference/query split over the evaluation set. 
    
    Evaluation consists of one pass over the reference images followed by one pass over the query images. During the reference pass, each category is split into manageable batches, and the class centroid is computed from a running total of the embedding vectors. The query pass is divided the same way, and the pre-computed centroids are used to make class predictions. When localization is used, we run 10 trials where different 10\% subsets of the reference images are randomly selected to receive bounding box annotations. 
    
    The above applies to prototype-styled classifiers, which require representation and query sets for a limited selection of classes in every batch. For straightforward softmax classifiers, we instead use random sampling without replacement, and a batch size of 128. Annealing schedule is as above. 

\end{document}